\documentclass[letterpaper, 10 pt, conference]{ieeeconf}  

\IEEEoverridecommandlockouts                              

\usepackage{graphics} 
\usepackage{epsfig} 
\usepackage{times} 
\usepackage{amsmath} 
\usepackage{amssymb}  

\usepackage{hyperref}
\usepackage{algorithm}
\newtheorem{problem}{Problem}
\newtheorem{condition}{Condition}
\usepackage{algpseudocode}
\usepackage{multirow}
\usepackage[table]{xcolor}
\usepackage{setspace}

\algnewcommand\algorithmicforeach{\textbf{for each}}
\algdef{S}[FOR]{ForEach}[1]{\algorithmicforeach\ #1\ \algorithmicdo}

\title{\LARGE \bf
Visual Action Planning with Multiple Heterogeneous Agents
}

\author{Martina Lippi*$^{1}$, Michael C. Welle*$^{2}$, Marco Moletta$^{2}$, Alessandro Marino$^{3}$, Andrea Gasparri$^{1}$, Danica Kragic$^{2}$
\thanks{*These authors contributed equally (listed in alphabetical order).}
\thanks{ ${}^1$Roma Tre University, Rome, Italy  {\it\small \{martina.lippi,andrea.gasparri\}@uniroma3.it} }%
\thanks{ ${}^2$KTH Royal Institute of Technology Stockholm, Sweden, {\it\small \{mwelle,moletta,dani\}@kth.se}}%
\thanks{ ${}^3$University of Cassino and Southern Lazio, Cassino, Italy {\it\small al.marino@unicas.it}}%
}

\begin{document}

\maketitle
\thispagestyle{empty}
\pagestyle{empty}

\begin{abstract}

Visual planning methods are promising to handle complex settings where extracting the system state is challenging. However, none of the existing works tackles the case of multiple heterogeneous agents which are characterized by different capabilities and/or embodiment. In this work, we propose a method to realize visual  action planning in multi-agent settings by exploiting a  roadmap built in a low-dimensional structured latent space and used for planning. To enable multi-agent settings, we infer possible parallel actions from a dataset composed of tuples associated with individual actions. Next, we evaluate feasibility and cost of them based on the capabilities of the multi-agent system and endow the roadmap with this information, building a capability latent space roadmap (C-LSR). Additionally, a capability suggestion strategy is designed to inform the human operator about possible missing capabilities when no paths are found. 
The approach is validated in a simulated burger cooking task and a real-world  box packing task.  
\end{abstract}


\section{Introduction and Related Work}

 Planning from raw observation~\cite{wang2019learning}, like images,  has proven very relevant in complex scenarios, such as when the scenes are highly dynamic and unstructured, as it eliminates the necessity to explicitly identify the system state.
Moreover, the use of raw observations paves the way for realizing visual action planning, i.e., for generating \emph{visual} plans, along with action plans, which allow to reach desired observations given start ones. The availability of visual plans also enhances the comprehension of the robot's plan by humans.

Several  prior works in the literature have investigated visual action planning methods. For instance, approaches operating in a high dimensional image space have been proposed in \cite{finn2017deep}, where a video prediction model is trained by resorting to a Long-Short Term Memory architecture and integrated into a Model Predictive Control formulation, and in  \cite{pmlr-v119-liu20h} where a graph structure is built from image sequence data.  More recently, many works have explored the possibility of mapping high-dimensional raw observations into lower-dimensional latent spaces to facilitate planning. For example,  a visual foresight module exploiting both RGB and depth data is proposed in~\cite{hoque2022visuospatial} for sequential fabric manipulation.  A structured latent space is learned and a graph-based roadmap is built in this space to perform planning in our previous works~\cite{lippi2022enabling,ace}.  Transformer models are adopted in \cite{sun2022plate} to realize visual planning from instructional videos, while an imitation learning method based on transporters with
visual foresight is proposed in \cite{wu2022transporters}.  

\begin{figure}[t]
    \centering
    \includegraphics[width=0.9\linewidth]{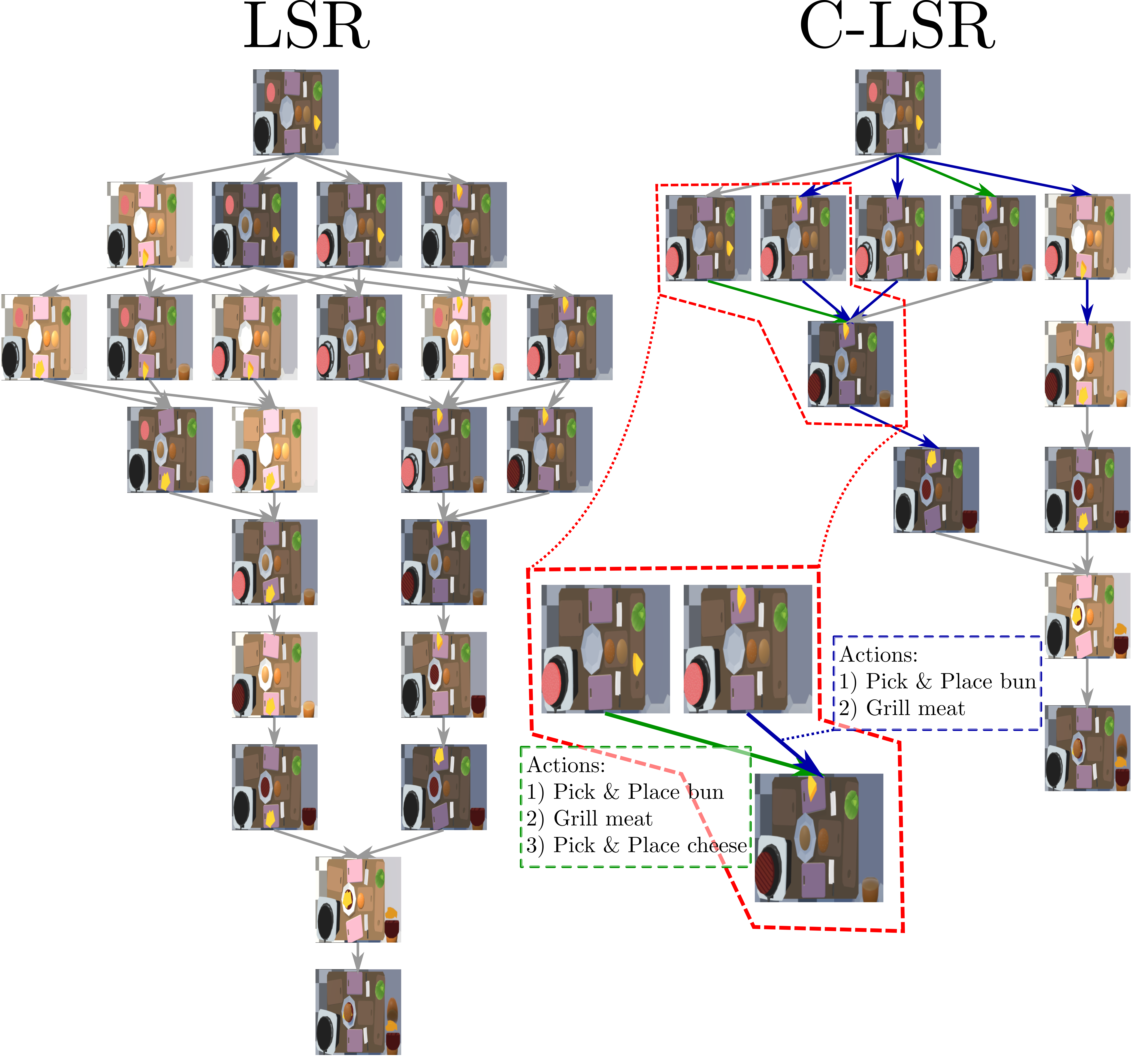}
    \vspace{-10pt}
    \caption{Example of VAPs obtained with an LSR (on the left), assuming one agent performing sequential actions, and with a C-LSR (on the right), enabling the parallel execution of multiple actions in multi-agent settings considering the agents' capabilities. Same start and goal configurations in a burger cooking task are used.  }
    \label{fig:fig1}
    \vspace{-10pt}
\end{figure}
However, all the above methods 
only consider a single agent that has to execute the action plan. In many scenarios, the availability of multiple \emph{heterogeneous} agents - meaning agents that have different capabilities and/or embodiment - is beneficial, if not essential, to successfully accomplish a given task. More specifically, when multiple agents are involved, their parallelism can be leveraged to reduce the overall execution time. Additionally, their diverse skills may prove crucial in completing all actions required for a given task, i.e., a single robot might not possess all the required skills to accomplish a task, and collaboration with other agents might be necessary. 
This is especially evident in multi-agent systems involving both human operators and robotic platforms. Indeed, many human cognitive and dexterous manipulation skills are still beyond the capability of even the most advanced robots.

Although many contributions exist in the literature that address multi-agent planning and allocation problems \cite{torreno2017cooperative,khamis2015multi}, these \emph{i)} do not provide visual information and \emph{ii)} often require extensive data regarding the actions to execute, which might not be easily retrieved. For instance, in multi-agent allocation and scheduling scenarios, it is frequently assumed that the set of actions to be performed is predefined, and precedence constraints are provided to specify whether an action must be executed before another, e.g.,~\cite{zhang2023energy,liu2023balanced,icra_ta}. Thus, in these cases, the primary task of the multi-agent strategy is to determine, for each action, which agent will execute it and the timing of its execution.

Differently from the state of the art, 
in this work, we propose a method to realize visual action planning with \emph{multiple heterogeneous} agents by relying on \emph{partial} data only.  This means that, given start and goal observations, the method must be capable of identifying the corresponding visual and action plans, as well as determining the assignment of actions to the available agents, taking into account their capabilities. Partial data availability is given by the fact that we only require a dataset composed of tuples collecting successor observations along with the information on the action that occurred between the two. This implies that no precedence constraints are provided. 
 We achieve the above by first building a Latent Space Roadmap (LSR)~\cite{lippi2022enabling} thought for a single agent, where each action has to be taken sequentially, and then extend it to the case of multiple agents that can possibly perform actions in parallel.
 In the first stage,  we assume that every action can be performed by an arbitrary agent and infinite agents with unlimited capabilities are available, and identify the set of actions that can be potentially executed in parallel in each state.  The resulting LSR, where edges associated with multiple parallel actions are introduced, is referred to as Parallel-LSR (P-LSR). 
In the second stage, starting from P-LSR, a Capability LSR is built where the given set of agents and their capabilities are taken into account. 
This enables us to plan paths from a given start to a goal state that respects the number and capabilities of the given agents while optimizing for relevant parameters of the multi-agent system, such as the overall workload and reachability of the agents. Fig. \ref{fig:fig1} shows an example of plans obtained using the LSR (on the left) and the C-LSR (on the right) obtained in a burger cooking task.  Given the same start (top) and goal (bottom) observations, the parallelism of multiple agents is exploited in the C-LSR to generate shorter paths. A zoom is provided (in red) to show examples of parallel actions with required skills. Assignments to agents are not shown for the sake of brevity.  

To the best of our knowledge, this is the first work enabling visual action planning with \emph{multiple heterogeneous} agents. 
In detail, our contributions are the following:
\begin{itemize}
    \item A novel algorithm to infer possible parallel actions from \emph{partial} data is designed. 
    \item A novel method is proposed to generate a \emph{capacity} latent space roadmap which enables visual action planning with multiple heterogeneous agents while taking into account their capabilities.
    \item A capability suggestion strategy is devised to inform human operators about possible missing capabilities in the multi-agent system to carry out the desired task. 
    \item Validation in simulated and real-world scenarios with heterogeneous multi-agent systems is provided. 
\end{itemize}

\section{Preliminaries}
In this section, we provide the preliminary notions for the proposed method. 
We refer to an action $u$ as an atomic operation  that is executed by a single agent, resulting in a change of the system state. For instance, actions can represent pick and place, pushing, screwing, welding, cutting, or stirring operations.  We denote with $\mathcal{U}$ the set of all feasible actions for the system at hand and with  $\mathcal{O}$ the space of possible system observations (e.g., images). 

\subsection{Dataset composition}\label{sec:dataset}
\begin{figure}
    \centering
    \includegraphics[width=0.7\linewidth]{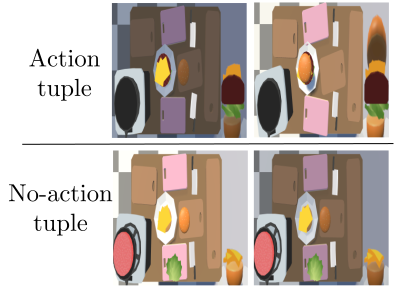}
    \vspace{-10pt}\caption{Example of action tuple ($b=1$) on the top and no action tuple ($b=0$) on the bottom for a burger cooking task. }
    \label{fig:dataset}
    \vspace{-10pt}
\end{figure}
 Similar to \cite{lippi2022enabling}, we assume that a dataset $\mathcal{T}_o$ is available. This is composed of tuples $(O_i,O_j,\rho)$ where $O_i\in\mathcal{O}$ and $O_j\in\mathcal{O}$ are two observations, and $\rho$ represents the action information between them. More specifically, this is defined as $\rho=(b,u)$, where $b\in \{0,1\}$ is  a binary indicator variable  denoting whether an action occurred ($b=1$) or not ($b=0$) between the two observations, and $u\in\mathcal{U}$ represents the respective action specification when $b=1$. 
\emph{No-actions} tuples (i.e., with $b=0$) capture the presence of task-irrelevant factors of variation in the observations~\cite{ccpaper}. For instance, variations in lighting conditions or alterations in the background may result in diverse observations  $O_i$ and $O_j$, which however are associated with the same underlying system state.  In these tuples, the value of the variable $u$ is ignored. Figure~\ref{fig:dataset} shows an action tuple ($b=1$)  and a no action tuple ($b=0$)  for a burger cooking task.

It is important to note that this dataset structure  \emph{i)} does not necessitate any knowledge on the underlying system states within the observations; rather, it solely requires information on the actions occurring between them; \emph{ii)} does not require to record full sequences of actions, with respective observations, but only relies on \emph{individual actions} performed by individual agents; and \emph{iii)} is agent-agnostic. These features contribute to a streamlined data collection process.

\subsection{Visual action planning in single-agent settings}\label{sec:seq-VAP}
The objective of visual action planning is to define, given the start $O_s$ and goal $O_g$ observations of the system, the sequence of actions to reach the goal and the respective sequence of observations. 
More formally, in a single-agent scenario, 
a \emph{sequential} Visual Action Plan (VAP) is defined as~\cite{lippi2022enabling}   \mbox{$P^{seq}=(P^{seq}_o,P^{seq}_u)$}, where  \mbox{$P^{seq}_o=(O_s=O_1,O_2,\cdots,O_N=O_g)$} represents a visual plan, containing a sequence of $N$ observations representing intermediate states from start to goal, and  \mbox{$P^{seq}_u=(u_1,u_2,...,u_{N-1})$} represents a sequential action plan, providing the respective actions $u_i$ to transition from $O_i$ to $O_{i+1}$, \mbox{$\forall i\in\{1,...,N-1\}$}.

\subsection{Latent Space Roadmap framework}

The Latent Space Roadmap (LSR) framework, proposed in our previous work~\cite{lippi2022enabling} and briefly recalled here, allows to realize sequential visual action planning in single agent scenarios. 
Briefly, this framework is based on mapping the high dimensional observations  in a lower dimensional \emph{structured} latent space  $\mathcal{Z}$ and then build a roadmap in this space to perform planning. 

\noindent \textbf{Latent space structure:}
We refer to the function mapping observations to a low dimensional structured latent space as latent mapping function $\xi:\mathcal{O} \to \mathcal{Z}$, and to the function performing the opposite mapping, i.e., from latent space to observation space, as observation generator function \mbox{$\omega:\mathcal{Z}\to \mathcal{O}$}. Ideally, the latent mapping function should be able to capture the underlying states of the system and structure the latent space accordingly. An approximated latent space is obtained using the dataset $\mathcal{T}_o$ described in Sec. \ref{sec:dataset} and resorting to an encoder-decoder architecture for modeling the functions $\xi$ and $\omega$. More specifically, a 
contrastive loss is exploited to structure the latent space: observations of no-action tuples (with $b=0$) are attracted, and observations of action tuples (with $b=1$) are repelled from each other. 
This enables to cluster together the same underlying states in $\mathcal{Z}$.

\noindent \textbf{Latent space roadmap:}
Based on the above clusters, an LSR is built which connects clusters based on the actions in $\mathcal{T}_o$. 
More specifically, an LSR    is a directed graph  \mbox{$\mathcal{G}=\{\mathcal{V},\mathcal{E}\}$} built in the latent space where each node in the set $\mathcal{V}$ is associated with a cluster of latent states (ideally corresponding to an underlying state of the system), while each edge $e=(i,j)$ in the set $\mathcal{E}$ is associated with a possible action $u_e$ to transition from node $i$ to node $j$  (according to action tuples in the dataset $\mathcal{T}_o$). In detail, let $z_i$ and $z_j$ be representative states of the two nodes and let $f:\mathcal{Z}\times \mathcal{U}\rightarrow \mathcal{Z}$ represent a transition function, then it holds $z_j=f(z_i, u_e)$.
This roadmap is used to find plans in the latent space given the latent encodings of start and goal observations. Plans in the latent space are  subsequently used to \emph{i)} generate action plans, by collecting the actions associated with the respective edges in the latent plan, and \emph{ii)}  generate visual plans, by decoding the latent states of the plan through the function $\omega$.

\section{Problem setting}

The scenario under consideration involves multiple heterogeneous agents, which can be robotic or human,  and are characterized by diverse skills. 
More specifically, let $\mathcal{A}=\{a_1,...,a_{n_a}\}$ be the set of $n_a$ available agents. 
For each agent $a_i\in\mathcal{A}$,
we define the following parameters: 
\begin{itemize}
    \item Set $\mathcal{S}^a_i$ of available skills, e.g., tools, sensors, or general abilities. 
    \item Average normalized workload $w_{i,j}\in[0,1]$ for performing the action $u_j$, $\forall u_j\in\mathcal{U}$. This can be dependent on physical properties, e.g., weight, or cognitive properties, e.g., fatigue in the case of human operators. 
\end{itemize}
In addition, for each agent we assume that a reachability function $r_i(x)\in[0,1]$ is available which assesses the feasibility and ease of reaching (and operating in) a specific pose $x$ by taking into account the agent kinematics and physical limits. In particular, we consider that $r_i(x)=0$ denotes the inability of the agent to reach the desired pose, whereas $r_i(x) = 1$ signifies that it can easily reach the pose. Examples of reachability indices can be found in~\cite{zacharias2007capturing,makhal2018reuleaux}.

Furthermore, for each action $u_j\in\mathcal{U}$, the following set of parameters is identified: 
\begin{itemize}
    \item Set $\mathcal{S}^u_j$ of skills, e.g., tools or sensors, that are required to perform the action.
    \item Set $\mathcal{P}_j$ of  relevant poses for the action which must be traversed to execute it. For instance,  in the context of pick and place actions, the pick and place poses could be included in $\mathcal{P}_j$. Similarly, for screwing actions, the screwing pose would be relevant to include in the set.
\end{itemize}
We define that an agent $a_i$ possesses the \emph{capability} to carry out an action $u_j$ if it/they has/have all the necessary skills for executing the action, i.e., $\mathcal{S}^u_j\subseteq \mathcal{S}^a_i$, as well as can reach all the respective relevant poses, i.e., $r_i(x_j)>0$, $\forall x_j \in\mathcal{P}_j$. We define an \emph{assignment} couple as $(a_i, u_j)$,  meaning that the agent $a_i$ is tasked with executing the action $u_j$.  Assignment couples must be \emph{valid}, indicating that the assigned actions align with the capabilities of the respective agents.
 In general, certain actions have the potential to be executed concurrently if multiple agents possessing the necessary capabilities are available. To identify these actions we introduce the following condition. 
\noindent
\begin{condition}\label{defn::pa}
Multiple actions $\{u_1, ... , u_p\}$ can be executed in parallel  if 
executing them in arbitrary order from a certain state results in the same final state. 
\end{condition}
For instance, given  two actions $u_i$ and $u_j$, these can be carried out in parallel from the state $z_k$
if it holds \mbox{$f(z_i, u_j) = f(z_j, u_i)$} where $z_i = f(z_k, u_i)$ and \mbox{$z_j = f(z_k, u_j)$}.
The rationale behind Condition \ref{defn::pa} is that, if the execution order does not matter, then, no precedence constraints (i.e., expressing actions that must be executed before/after others)  between the actions exist and  these can be carried out concurrently. Note that we do not take into account potential space constraints that may arise when executing actions in parallel, but we  assume that a low-level motion planner is available which prevents collisions between the agents. 

Based on the above, 
we can expand the definition of sequential VAP given in Sec. \ref{sec:seq-VAP} to a multi-agent setting. More in detail, we can define  a \emph{parallel} VAP \mbox{$P^{par}=(P^{par}_o,P^{par}_u)$} where, in contrast to $P^{seq}$, the action plan $P^{par}_u$ enables the execution of multiple actions in parallel by different agents. This is defined as \mbox{$P^{par}_u=(\overline{\mathcal{U}}_1,\overline{\mathcal{U}}_2,...,\overline{\mathcal{U}}_{N-1})$}, where $\overline{\mathcal{U}}_k$ represents the collection of \emph{assignment} couples $(a_i,u_j)$, denoting that the agent $a_i$ has to execute action $u_j$. All the actions in each set $\overline{\mathcal{U}}_k$ are executed in parallel. The visual plan \mbox{$P^{par}_o=(O_s=O_1,O_2,\cdots,O_N=O_g)$} collects the  sequence of $N$ observations which are obtained by applying the (possibly parallel) actions in $\overline{\mathcal{U}}_i$.

We can now state the main problem addressed in this work.

\begin{problem}\label{prob}
    Consider a heterogeneous multi-agent system with set of agents $\mathcal{A}$. Assume a dataset $\mathcal{T}_o$ is available and start $O_s$ and goal $O_g$ observations are assigned. Our objective is to generate parallel VAPs, $P^{par} = (P^{par}_o, P^{par}_u)$, such that \emph{i)} they provide visual and action plans to reach the goal state, \emph{ii)}    the assignment couples are  \emph{valid},  and \emph{iii)} the overall workload and reachability indices are optimized.  
\end{problem}

\section{Visual Action Planning in Heterogeneous Multi-agent Systems}
In this section, we present the proposed framework for addressing the above problem. 

\subsection{Solution overview}
Our core idea is to  infer all the possible actions that can be executed in parallel by exploiting the dataset $\mathcal{T}_o$  and the respective LSR framework, and subsequently build a new  roadmap in the latent space  that incorporates these actions, enabling planning with multiple agents. 
In doing so, the agents capabilities and the actions requirements are taken into account. 
More in detail, we resort to Condition~\ref{defn::pa} to identify potential parallel actions and define a Parallel LSR (P-LSR). This represents a directed graph $G^{par} = (\mathcal{V}, \mathcal{E}^{par})$ where the set of edges encodes potentially parallel actions that are executable by a multi-agent system, \emph{regardless} of the number of agents and their individual capabilities. Hence, each  edge $e = (i,j)$ in the set $\mathcal{E}^{par}$ is associated with a set of actions $\mathcal{U}_e$, all of which must be executed to transition from node $i$ to node $j$.
The set of nodes coincides with the LSR one, i.e., it collects the latent space clusters associated with different underlying system states. 

Next, we build a \emph{capability} LSR, denoted as C-LSR, that takes into  account the agents capabilities and the actions requirements. This is defined as a directed graph \mbox{$G^{c} = (\mathcal{V}, \mathcal{E}^{c})$} where the set of edges encodes possible assignment couples by considering the agents at hand. More specifically, each  edge $e = (i,j)$ in the set $\mathcal{E}^{c}$ is associated with a set $\overline{\mathcal{U}}_e$ of \emph{valid} assignment couples and with a \emph{cost} $c_e$, quantifying the effectiveness of the multi-agent system to perform the actions in $\overline{\mathcal{U}}_e$. 
Similar to the above, the set of nodes remains unchanged with respect to the LSR. 
This graph is used online to generate parallel visual action plans given start and goal observations. Additionally, we define a \emph{capability suggestion} strategy which, in situations where no plan is found, proposes  to the human operator the capabilities that are missing in the multi-agent system to reach the desired goal state.

In the following, we first detail  the procedure for identifying actions in an LSR that can be executed in parallel and explain how to incorporate these new edges to form the \mbox{P-LSR}. Then, we present the method to build the C-LSR that takes into account a given set of agents and their capabilities. Finally,  we outline the online visual action planning strategy along with the capability suggestion method.

It is worth noticing that the dataset $\mathcal{T}_o$ does not specify any explicit dependency between the actions, e.g., no precedence constraints  are  defined. Similarly, no information about actions that can be potentially executed concurrently is provided. Instead, it comprises simple individual actions collected without consideration of multi-agent settings.

\subsection{Parallel LSR (P-LSR)}
In order to obtain P-LSR from the LSR $\mathcal{G}$, we leverage Condition \ref{defn::pa}, i.e., actions are executable parallel if their execution in arbitrary order yields the same results.
 We define the set of actions that can be executed in a certain node $n$ as the collection of the actions associated with all the edges originating from $n$.  
The basic idea of our algorithm is that if all actions associated with a certain path can be executed starting from
the first node of the same path,  then these actions can be executed in parallel.
For example, let us assume that the actions associated with a path between the nodes $n$ and $t$ are $\mathcal{U}_{nt}=\{u_1,u_2\}$  and the set of actions that can be executed from   node $n$ is $\mathcal{U}_n=\{u_1,u_2,u_3\}$, then the set of actions that can be executed in parallel from node $n$ to node $t$ is $\mathcal{U}_p=\{u_1,u_2\}$.
 In detail, Algorithm \ref{alg::plsr} shows how the above logic is applied from every node in $\mathcal{V}$ to every other node in $\mathcal{V}$. 
 
\begin{algorithm} \caption{P-LSR building}
\small
\def\negsp{\vspace{-5pt}}
\def\negup{\vspace{-7pt}}
\def\negin{\vspace{-3pt}}
\setstretch{1.0}
\begin{algorithmic}[section]
\Require LSR $\mathcal{G}=(\mathcal{V},\mathcal{E})$, threshold $\tau$
    \begin{algorithmic}[1]
    \State $\mathcal{V}^{par}=\mathcal{V}$
    \State $\mathcal{E}^{par}=\mathcal{E}$
   \ForEach{$n \in \mathcal{V}$, $t \in \mathcal{V}, n\neq t$}
            \If{has-path-longer-one$(\mathcal{G},n,t)$} \label{pa:line:haspath}
                \State $\mathcal{U}_n=$get-all-actions-from-node$(n)$\label{pa:line:act}
                \State $SP_{nt}=$all-shortest-paths$(\mathcal{G},n,t)$\label{pa:line:allshortestp} 
                \ForEach{$P_{nt} \in SP_{nt}$}
                    \State $\mathcal{U}_{nt}=$get-path-actions$(P_{nt})$ \label{pa:line:getpathedges}
                    \State $\mathcal{U}_p=$compute-intersection$(\mathcal{U}_n,\mathcal{U}_{nt},\tau)$ \label{pa:line:getinter}
                    \If{$|\mathcal{U}_{nt}| = |\mathcal{U}_p|$}
                        \State $\mathcal{E}^{par} \leftarrow $ add-edge$(\mathcal{U}_p)$\label{pa:line:edge}
                    \EndIf
                \EndFor
            \EndIf       
        
    \EndFor
   
    \end{algorithmic}
\Return $\mathcal{G}^{par}=(\mathcal{V}^{par},\mathcal{E}^{par})$
\end{algorithmic}
\label{alg::plsr}
\vspace{-1pt}
\end{algorithm}
 
 At the beginning, the P-LSR sets $\mathcal{V}^{par}$ and $\mathcal{E}^{par}$ are  initialized as the LSR sets $\mathcal{V}$ and $\mathcal{E}$, respectively. Then, for each couple of nodes  $n$ and $t$,  we check if there exists a path from $n$ to $t$ with a minimum length of two (line \ref{pa:line:haspath}).  If so, we compute the action set $\mathcal{U}_n$  containing all actions that can be executed in node $n$ (line \ref{pa:line:act}) and
 find all the shortest paths $SP_{nt}$ from node $n$ to node $t$ (line \ref{pa:line:allshortestp}).  For each shortest path $P_{nt}$, we extract the respective set of actions corresponding to the edges in the path and denote it as $\mathcal{U}_{nt}$.
 
 At this point, 
 the intersection set $\mathcal{U}_p$ between $\mathcal{U}_n$ and $\mathcal{U}_{nt}$ is calculated to identify actions that can be executed in parallel from node $n$ (line \ref{pa:line:getinter}). This intersection is computed by considering that two actions $u_n \in \mathcal{U}_n$ and $u_{nt} \in \mathcal{U}_{nt}$  can be assumed equivalent if \emph{i)} the respective sets of required skills coincide,  i.e. $\mathcal{S}^u_n = \mathcal{S}^u_{nt}$, and 
 \emph{ii)}  the (Euclidean) distances between the respective relevant action poses in $\mathcal{P}_n$ and $\mathcal{P}_{nt}$ are below a threshold $\tau$.
Based on this intersection, a new edge, with set of parallel actions $\mathcal{U}_p$, is added to $\mathcal{E}^{par}$ (line \ref{pa:line:edge}) if $\mathcal{U}_p$ contains the same number of actions as the set $\mathcal{U}_{nt}$,   thus guaranteeing that all actions in the path $\mathcal{P}_{nt}$ are executable from node $n$ and can be then carried out concurrently. 

Note that Algorithm \ref{alg::plsr} is agent-agnostic, meaning it does not require the set of agents to be specified, but builds all possible edges into $\mathcal{G}^{par}$ that fulfill Condition \ref{defn::pa} given $\mathcal{G}$. This enables us to reuse Algorithm \ref{alg::plsr} in case the set of available agents changes.

\subsection{Capability matching and capability LSR (C-LSR)}
 Our objective is now to establish whether the heterogeneous multi-agent system possesses the capabilities to perform the parallel actions associated with the edges in $\mathcal{E}^{par}$, along with the respective \emph{cost} to execute them.  In this way, given the P-LSR, we can generate the capability LSR.

We first introduce the cost $c_{ij}$ for the agent $i$ to perform the action~$j$ by taking into account 
    \emph{i)} the agent reachability function,  
    \emph{ii)} the agent workload performing the action, and \emph{iii)} the agent availability of needed tools/sensors. More specifically,
    the cost is obtained as
\begin{equation}\label{eq:cij}
c_{i,j} \!= \!\left\{
\begin{aligned}
 &\alpha \frac{1}{|\mathcal{P}_j|} \sum_{k\in\mathcal{P}_j} \left(1-r_i(x_k)\right) + \beta w_{i,j}, &&\!\!\!\text{if} \,\,
 \begin{aligned}[t]
     &\mathcal{S}^u_j\subseteq\mathcal{S}^a_i,\text{and} \\ &\!\!\!\!r_i(x_j)>0, \,\forall x_j,
 \end{aligned} 
 \\
 &\infty & &\text{otherwise}
\end{aligned}    
\right.
\end{equation}
with $\alpha$ and $\beta$ positive constants and $|\mathcal{P}_j|$ the cardinality of the set $\mathcal{P}_j$.
 This implies that the condition $c_{ij}<\infty$ denotes a valid assignment couple $(a_i, u_j)$, while the case $c_{ij}=\infty$ indicates a couple that is not valid, 
 meaning that the agent does not possess the necessary equipment and/or cannot reach all the desired action poses. 
Based on the above, the C-LSR is constructed as outlined in Algorithm \ref{alg::weighplsr}. 

\begin{algorithm} \caption{Capability LSR building}
\small
\def\negsp{\vspace{-5pt}}
\def\negup{\vspace{-7pt}}
\def\negin{\vspace{-3pt}}
\setstretch{1.0}
\begin{algorithmic}[section]
\Require P-LSR $\mathcal{G}^{par}=(\mathcal{V}^{par},\mathcal{E}^{par})$, Agents $\mathcal{A}$
    \begin{algorithmic}[1]
    \State $\mathcal{V}^{c}=\mathcal{V}^{par}$\label{alg:wplsr:nodes}
    \State $\mathcal{E}^{c}=\{\}$
   \ForEach{$e \in \mathcal{E}^{par}$}
        \State $\mathcal{U}_e=$get-edge-actions$(n)$\label{alg:wplsr:getactions}
        \ForEach{$a_i \in\mathcal{A}$, $u_j\in\mathcal{U}_e$}
           \State $c_{i,j} \leftarrow$ compute-cost($a_i,u_j$) [Eq.~\eqref{eq:cij}]
        \EndFor
        \State $X \leftarrow $ solve-ILP-assignment($\mathcal{U}_e, \mathcal{A},c$)\label{alg:wplsr:ILP}
        \If{$X$ feasible and finite objective} 
                \State $\overline{\mathcal{U}}_e \leftarrow $ get-assignment-couples($X$) \label{alg:wplsr:assign}
                \State $c_e$ compute-edge-cost($\mathcal{U}_e, \mathcal{A},c,X$) [Eq. \eqref{eq:cost}]\label{alg:wplsr:cost}
                    \State $\mathcal{E}^{c} \leftarrow $ add-edge$(\overline{\mathcal{U}}_e , c_e)$
                \EndIf 
        
    \EndFor
   
    \end{algorithmic}
\Return $\mathcal{G}^{par}=(\mathcal{V}^{par},\mathcal{E}^{par})$
\end{algorithmic}
\label{alg::weighplsr}
\vspace{-1pt}
\end{algorithm}

First, the set of nodes is initialized starting from the P-LSR one (line \ref{alg:wplsr:nodes}). Next, for each edge in $\mathcal{E}^{par}$ the respective set of $n_e$ possible parallel actions $\mathcal{U}_e=\{u_{e,1},..,u_{e,n_e}\}$ (line \ref{alg:wplsr:getactions}) is analyzed. In particular, we have to establish: whether the actions can be executed in parallel by the available agents $\mathcal{A}$, and, if so, how actions should be optimally allocated to the agents. 
To this aim, let us introduce the binary decision variable  $X_{i,j}\in\{0,1\}$, $\forall i\in\mathcal{A}, j\in\mathcal{U}_e$, which is $1$ if the agent $i$ is assigned to the action $j$ and is $0$ otherwise.  To determine the decision variables, we formulate a simple Integer Linear Programming (ILP) problem  as follows (line \ref{alg:wplsr:ILP} where $c$ and $X$ denote the collective cost and decision variables, respectively)
\begin{subequations}\label{eq:opt}
\begin{align}
    &\min_{X_{i,j}} && \sum_{i\in\mathcal{A}}\sum_{j\in\mathcal{U}_e} c_{i,j} X_{i,j} \label{eq:opt:cost}
    \\ & \text{s.t.} && \sum_{i\in\mathcal{A}} X_{i,j} = 1, \quad \forall j\in\mathcal{U}_e \label{eq:opt:task}
    \\ & && \sum_{j\in\mathcal{U}_e} X_{i,j} \leq 1, \quad \forall i\in\mathcal{A}. \label{eq:opt:agent}
\end{align}
\end{subequations}
More specifically, the objective function in \eqref{eq:opt:cost} aims to minimize the overall cost to carry out the actions in $\mathcal{U}_e$, the equality in \eqref{eq:opt:task} ensures that all actions are executed by an agent, and the inequality in \eqref{eq:opt:agent} guarantees that each agent can carry out at most one action.

In case the above problem is unfeasible or results in an objective function with infinite value, it means that the available agents lack the capability to fulfill the actions in $\mathcal{U}_e$ and no edge associated with $\mathcal{U}_e$ is added to the set of edges $\mathcal{E}^{c}$.
In contrast, if the problem is feasible and leads to a finite objective function, 
an edge $e$ is added to $\mathcal{E}^{c}$, where the assignment couples $\overline{\mathcal{U}}_e$ are derived from the decision variables $X_{i,j}$, $\forall i\in\mathcal{A}, j\in\mathcal{U}_e$ (line \ref{alg:wplsr:assign}), while the cost (line \ref{alg:wplsr:cost}) is obtained as  
\begin{equation}\label{eq:cost}
    c_e = \gamma\sum_{i\in\mathcal{A}}\sum_{j\in\mathcal{U}_e} c_{i,j} X_{i,j}+ \mu \frac{1}{|\mathcal{U}_e|},
\end{equation}
with $\gamma$ and $\mu$ positive weights. Hence, the cost is composed of a first contribution given by the objective function resulting from \eqref{eq:opt} and a second contribution related to the number of parallel actions, i.e., the higher the number of parallel actions and the lower the contribution, thus encouraging plans which maximize the possible parallelism in the system.

It is worth noticing that both P-LSR and C-LSR are built offline, and can be used at runtime for planning purposes given arbitrary start and goal observations. Moreover, as mentioned in the previous section, if the set of agents $\mathcal{A}$ varies, only C-LSR must be recalculated.

\subsection{Online visual action planning}\label{sec:suggestion}
Given the capability LSR, parallel VAPs fulfilling the requirements in Problem~\ref{prob} can be easily found. More in detail, given the start and goal observations $O_s$ and $O_g$, respectively, these are mapped into latent states $z_s$ and $z_g$ through the latent mapping function $\xi$. Then, the respective closest nodes in the C-LSR are retrieved and Dijkstra's algorithm is applied to find the latent path from start node to goal one with minimum overall cost. The visual plan is then obtained by decoding (through the function $\omega$) the latent states associated with the nodes in the latent path, while the parallel action plan is obtained by considering the assignment couples in the edges of the latent path. 

\noindent
\textbf{Capability suggestion:}
 In case no path is found, this might indicate that some required capabilities to carry out the actions to reach the goal state are missing within the multi-agent system. Hence, a capability suggestion strategy is designed to inform the human operator about missing capabilities that might be needed to successfully execute the task. 
 To this aim, we resort to the LSR $\mathcal{G}$ where no capabilities of the agents and parallel actions are taken into account. More in detail, we retrieve the closest nodes in $\mathcal{V}$ with respect to $z_s$ and $z_g$ and find the shortest path in $\mathcal{G}$. 
 If such a path is found, it indicates that the multi-agent team is missing a certain capability to reach the goal.
Hence, for each action $u_j$ in the plan (retrieved from the edges), the capabilities of the multi-agent system to perform $u_j$ are evaluated, i.e., the cost $c_{i,j}$ in \eqref{eq:cij} is computed for each  $i\in\mathcal{A}$. If it holds $c_{i,j} = \infty$ for all agents, then no agent is able to to perform the action. 
In this case, for each agent $j$, we return: 
 \begin{itemize}
     \item The set of skills that are required for the action and are not available for the agent, i.e., $\mathcal{S}^u_j \setminus (\mathcal{S}^u_j \cap \mathcal{S}^a_i)$. 
     \item The set of relevant poses of the action which are not reachable, i.e., such that $r_i(x_j)=0$ with $x_j\in\mathcal{P}_j$. 
 \end{itemize} 
 Furthermore, the decoded observations associated with the nodes connected by the analyzed edge are returned to provide a visual understanding to the human operator. 
With access to this information, the human operator can gain insights into the system's missing capabilities and potentially integrate them in an informed manner.

\section{Simulation results}\label{sec:exp}
\begin{figure}
    \centering
    \includegraphics[width=0.45\linewidth]{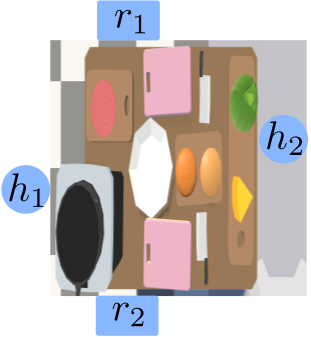}
    \vspace{-5pt}
    \caption{Simulated burger cooking setup. Blue rectangles denote the position of the bases of two robots $r_1,r_2$, while blue circles denote the starting positions of two human operators $h_1,h_2$. }
    \label{fig:setup-sim}
    \vspace{-10pt}
\end{figure}

In this section, we showcase the performance of the proposed framework on a burger cooking task simulated using Unity3D engine \cite{unitygameengine}, as depicted in Figs~\ref{fig:dataset} and~\ref{fig:setup-sim}. 

\noindent \textbf{Setup description:}
Given the available ingredients, the burger is assembled on the white plate in the middle of the table. This requires the execution of a variety of actions, which need different skills to be executed.  More specifically, the objects involved in the scene are: meat patty, cheese, lettuce, and the top and bottom parts of the bun. All these objects can be moved within the cooking station through pick and place actions (requiring gripping skills). Moreover, cheese and lettuce can be sliced (requiring cutting skills), while meat patty can be cooked on the pan (requiring grilling skills).  The table size is $1.4\,\text{m}\times 0.8$~m, with height $0.9$~m. All the objects lie on the table, except for the  pan which is positioned at a height of $1.2$~m.  Given the cooking  task setting, a maximum of four actions can be executed in parallel. 
 A dataset composed of a total of $5000$ tuples was collected. This is divided in $58\%$ action tuples (with $b=1$,  Fig~\ref{fig:dataset} - top) and $42\%$ no-action tuples (with $b=0$,  Fig~\ref{fig:dataset} - bottom). 
Variations in lighting conditions and  scales of the objects as well as noise in the positioning of ingredients and the cutting boards on the table are simulated in the observations of the dataset. These factors of variation, although irrelevant for achieving the desired task~\cite{ccpaper}, are introduced to simulate realistic environmental conditions.
 
 As far as the set of agents is concerned, we consider that a maximum of four agents may be present in the scene, comprising two robotic units, denoted as $r_1$ and $r_2$, and two human operators, denoted as $h_1$ and $h_2$. Their positioning is illustrated in  Fig~\ref{fig:setup-sim},  with the robots' bases mounted at the table height. 
We assume that robotic agents possess gripping and cutting skills but do not have the grilling skill (needed for the meat patty). In contrast, human operators are able to perform all the actions. However, we assume that a higher workload is associated with human operators (equal to $1$ for all actions) compared to robots (where, for all actions,  workload equal to $0.5$ is assumed for $r_1$ and $0.3$ for $r_2$). A simple proportional rule between the target position and the agent base position is adopted for the reachability index. This distance is normalized with respect to the maximum distance attainable by the respective agent, which is set to $1.5$~m for robots and $5$~m for humans  resulting in the fact that the humans can reach every task station with ease.

The  LSR framework is built by following the settings of our previous work~\cite{lippi2022enabling}, where a Variational Auto-Encoder (VAE) was used to model the functions $\xi$ and $\omega$ and an optimization procedure was proposed to tune the clustering threshold. More in detail, we set the maximum number of weakly connected graph components equal to $1$ and bounded the clustering threshold in the range $[0,3]$. 
Regarding the parameters for the C-LSR, we choose weights $\alpha=\beta=\gamma=\mu=1$. 
For performance assessment, we select $1000$ different start and goal observations from a novel holdout set and evaluate the correctness of the generated parallel VAPs. 
\begin{figure}
    \centering
    \includegraphics[width=0.9\linewidth]{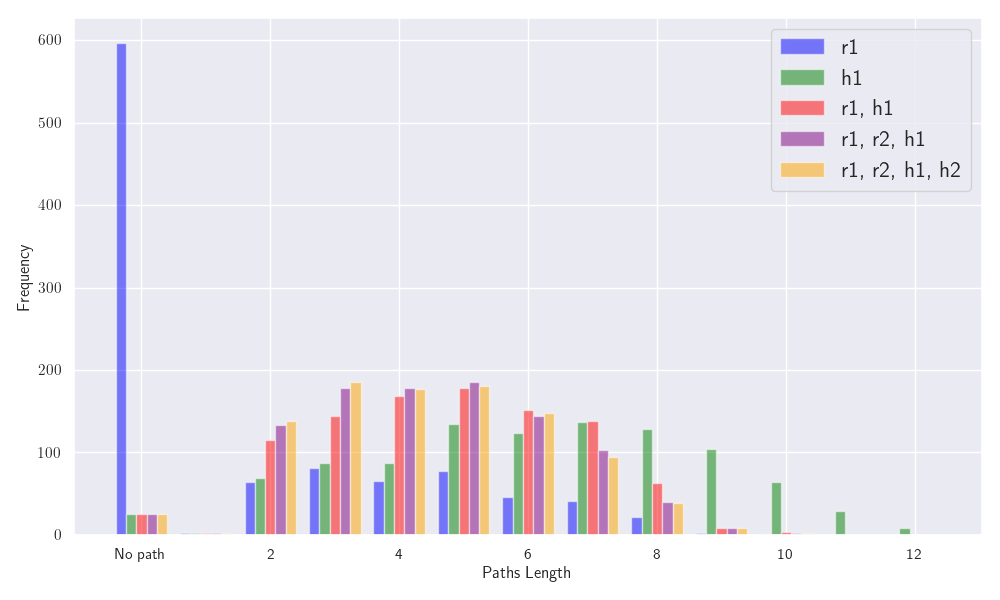}
    \vspace{-10pt}
    \caption{Histograms of the path lenghts $N$ with different sets of agents. }
    \label{fig:hist}
    \vspace{-10pt}
\end{figure}
\begin{figure*}
    \centering
    \includegraphics[width=0.9\linewidth]{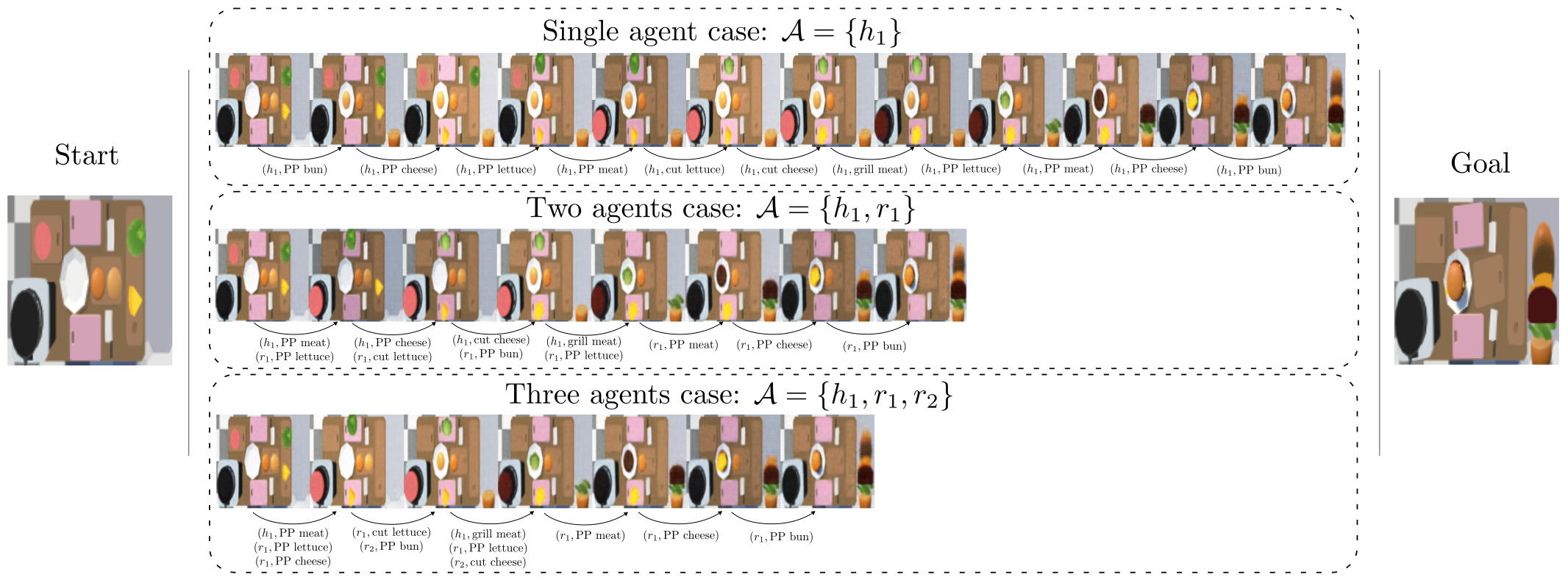}
    \vspace{-10pt}
    \caption{Examples of parallel VAPs obtained with different sets of agents given the same start (on the left) and goal (on the right) observations. }
    \label{fig:capability-sim}
    \vspace{-10pt}
\end{figure*}
\noindent
\textbf{Results with different sets of agents:} 
In the following,  the effectiveness of the proposed C-LSR with different sets of agents is validated. More in detail, we analyze the results with the following sets of agents: 
$\{r_1\}$, $\{h_1\}$, $\{r_1,h_1\}$, $\{r_1,r_2,h_1\}$, and $\{r_1,r_2,h_1,h_2\}$. 
In all cases, all actions selected as potential parallel actions are correctly combined and we obtain a percentage of correct individual transitions in the paths equal to $\approx 97\%$, and, except for \mbox{$\mathcal{A}=\{r_1\}$}, a percentage of full correct paths of $\approx 82\%$. Note that the latter percentage also takes into account the events when no path is found.  For the \mbox{$\mathcal{A}=\{r_1\}$} case, as detailed in the following, a lower percentage of correct paths, equal to $35\%$, is obtained as valid paths are frequently not available.  

Figure~\ref{fig:hist} reports the histograms of the path lengths $N$ obtained with the different sets of agents. More in detail, when only robot $1$ is included (shown in blue), no paths are found $597/1000$ times. This is motivated by the fact that the robot does not possess the grilling skill for the meat patty which is frequently required. In contrast, when only the human operator $1$ is included (in green), a path is obtained in the majority of cases, with only $26$ instances where it is not found. As this agent possesses all necessary skills and can access all objects, this scenario mirrors the performance of the simple LSR in single-agent settings. In this case, average path length equal to $\approx 6.3$ is obtained, while maximum path length of $12$ is observed.
When adding a robotic agent (in red), the generated paths lengths significantly reduce achieving average equal to $\approx 4.8$ and maximum equal to $10$. Finally, additional (smaller) improvements are  observed when expanding the set of agents further, reaching average $\approx 4.8$ and maximum equal to $9$ with the full set of agents. 

Figure~\ref{fig:capability-sim} shows examples of parallel VAPs obtained with different sets of agents given the same start (on the left) and goal (on the right) observations (PP denotes pick and place operations in the figure). The objective is to prepare the complete burger with bun,  cut lettuce, (grilled) meat, and slices of cheese. In the top row, the case of a single human agent is analyzed. In this case, all the actions are executed sequentially until the desired state is reached, leading to an overall workload of the human equal to $11$. When a robot is added to the team (second row), the parallelism between the two agents is exploited in the first actions of the plan, while fulfilling the validity of the assignments and minimizing the overall path cost. More in detail, the grilling operation is assigned to the human in step four of the path, while the final three assembly steps, which have to be executed sequentially, are assigned to the robot to minimize the human effort (recording an overall effort equal to $4$). Finally, introducing a second robotic unit to the team, as shown in the third row, further maximizes the exploitation of parallelism, resulting in an overall human workload of $2$. 

\begin{figure}
    \centering
    \includegraphics[width=0.7\linewidth]{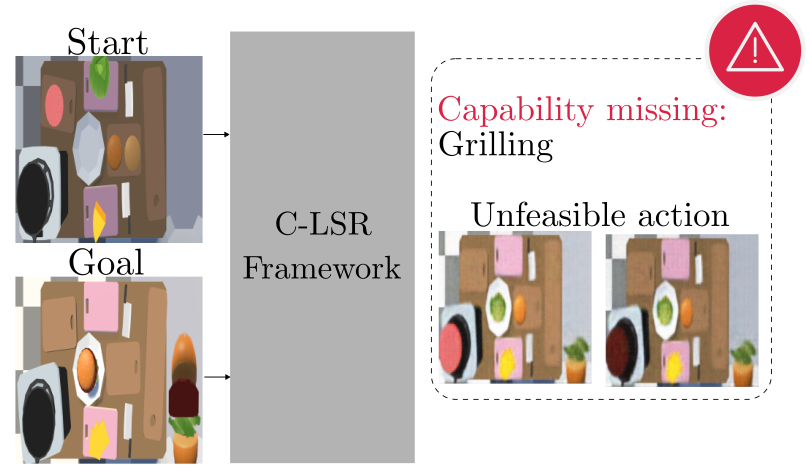}
    \vspace{-12pt}
    \caption{Example of outcome of the missing capability suggestion strategy.}
    \label{fig:capability-sugg-sim}
    \vspace{-15pt}
\end{figure}

\noindent \textbf{Missing capability suggestion strategy:} The strategy to inform human operators about missing capabilities has been validated with a team composed of robotic agents only, i.e., $\mathcal{A}=\{r_1,r_2\}$. An example is provided in Figure~\ref{fig:capability-sugg-sim}. More specifically, the agents are required to reach the goal observation  (having burger with meat and salad) in the bottom left starting from the observation in the top left. The proposed C-LSR framework is employed to find the parallel VAPs, but no paths are found. Hence, the strategy in Sec.~\ref{sec:suggestion} is executed to offer suggestions to the human operator regarding potential missing capabilities. This results in identifying that the grilling skill is missing within the current team (on the right). A visual representation of the unfeasible action is also provided for interpretability purposes.  

\section{Experimental results}
\begin{figure}
    \centering
    \includegraphics[width=0.8\linewidth]{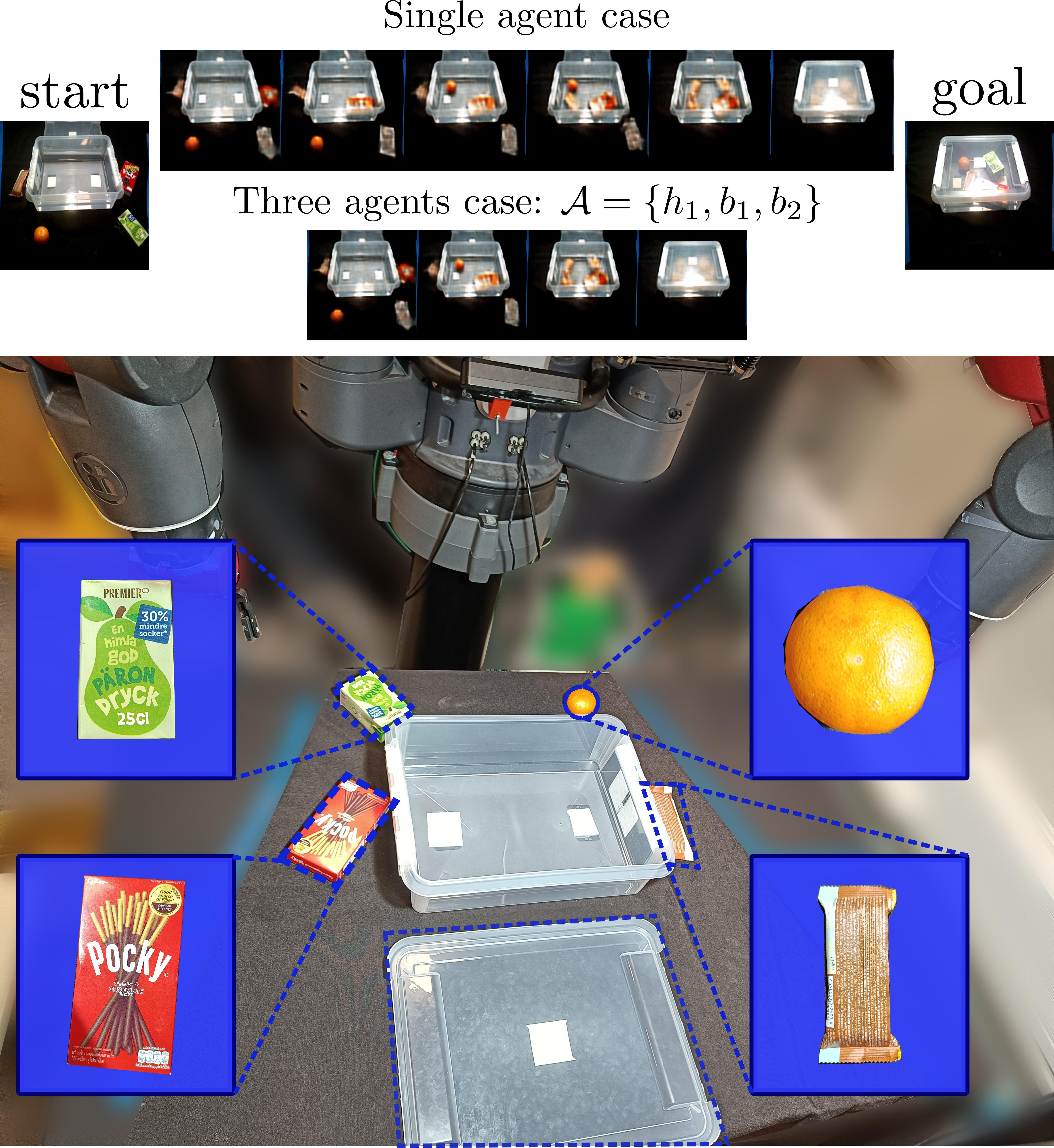}
    \vspace{-5pt}
    \caption{Depiction of the real-world packing task and the generated parallel VAPs with different sets of agents.}
    \label{fig:baxter}
    \vspace{-20pt}
\end{figure}
In this section, we present the experimental results of the C-LSR in a real-world box packing task, shown in Fig.~\ref{fig:baxter}, involving three heterogeneous agents. The objective is to pack (and possibly close) the box in the middle of the table. Five objects are involved in the task: juice box, mandarin, chocolate sticks,  granola bar, and box cover. 
The first four items can be moved within the box (requiring gripping skills), while the latter has to be placed on the box to close it (requiring dexterous manipulation skills).  A dataset composed of $900$ tuples (divided into $54\%$ action tuples and $46\%$ no-action tuples) was collected. Variations in the object positioning and the lighting conditions were naturally captured in the dataset.   
Regarding the agents, we consider two robotic arms $b_1,b_2$ (left and right arms of a Baxter robot) and a human operator $h_1$. The two arms have gripping skills only, while the human has dexterous manipulation skills. Similar to the simulated case study, for all actions, the workload is set to $1$ for the human and to $0.5$ for the two arms. For the other parameters of the framework, the same settings as for the simulated case study are used. 
The Baxter workspace limits result in the fact that only the two objects closest to each arm can be gripped by the respective arm and safely put into the box.
We validate the approach by requiring to fully pack the box (i.e., put the four food items in it) and close the cover, given a starting configuration (shown in Fig.~\ref{fig:baxter}) where all the food items are outside of the box and placed on the table. 
The top part of the figure shows the plans obtained with two different sets of agents. 
In the first line, a single agent capable of performing all the actions is assumed and all the steps to pack the box are executed sequentially. In this example, the items are moved in the following sequence: chocolate, mandarin, granola, juice, and box cover, leading to a path length equal to $6$.  
When the three agents are considered (second line), we can observe that the parallelism of the system is fully exploited. Indeed, in the first step, the right arm is required to move the chocolate box into the box, and, concurrently, the left arm is required to move the mandarin in the box. Next, the right arm is tasked with moving the juice box, while the left arm has to move the granola bar. Finally, the human is required to put the cover, completing the task, leading to a length of the path equal to $4$.   The accompanying video shows all the steps of the parallel execution in the box packing experiment. 
\section{Conclusion}
In this paper we proposed a visual action planning framework for heterogeneous multi-agent settings, which is suitable to involve human operators. Our method relies on partial data, where only tuples collecting observations of successor states with the respective action information are required. The proposed framework is based on first identifying actions that can be potentially executed in parallel considering an arbitrary number of agents with unlimited capabilities and then building a capability latent space roadmap  (C-LSR) that takes into account the set of available agents and their capability. A strategy to suggest missing capabilities to accomplish desired tasks was also proposed. We validated the effectiveness of our approach on simulated and real-world data. As future work, we aim to integrate multi-model foundation models to achieve a natural human-robot interaction.

\bibliographystyle{ieeetr}
\bibliography{references}

\end{document}